# Trustworthy, responsible, ethical AI in manufacturing and supply chains: synthesis and emerging research questions

Alexandra Brintrup[a*], George Baryannis[b], Ashutosh Tiwari[c], Svetan Ratchev[d], Giovanna Martínez-Arellano[d] and Jatinder Singh[e]

[a]Institute for Manufacturing, University of Cambridge, Cambridge, UK; [b]School of Computing and Engineering, University of Huddersfield, Huddersfield, UK; [c]University of Sheffield, Sheffield, UK; [d]University of Nottingham, Nottingham, UK; [e]Department of Computer Science and Technology, University of Cambridge, Cambridge, UK

ABSTRACT

While the increased use of AI in the manufacturing sector has been widely noted, there is little understanding on the risks that it may raise in a manufacturing organisation. Although various high level frameworks and definitions have been proposed to consolidate potential risks, practitioners struggle with understanding and implementing them. This lack of understanding exposes manufacturing to a multitude of risks, including the organisation, its workers, as well as suppliers and clients. In this paper, we explore and interpret the applicability of responsible, ethical, and trustworthy AI within the context of manufacturing. We then use a broadened adaptation of a machine learning lifecycle to discuss, through the use of illustrative examples, how each step may result in a given AI trustworthiness concern. We additionally propose a number of research questions to the manufacturing research community, in order to help guide future research so that the economic and societal benefits envisaged by AI in manufacturing are delivered safely and responsibly.

## 1. Introduction

Artificial Intelligence (AI) has been a major driver of Industry 4.0, with diverse and rich use cases that have helped improve productivity through efficiency gains. Manu-facturers increasingly seek AI based solutions for major challenges the sector is facing; from improving supply chain resilience (Hosseini and Ivanov, 2020), to achieving cli-mate and sustainability goals (Naz et al., 2022). In the UK, 68% of large companies, 34% of medium sized companies and 15% of small companies have adopted at least one AI technology, with 44% overall expressing an interest to adopt in the next three years (DCMS, 2022).

However, the manufacturing industry is also at a crossroads with AI technology. While trustworthy, responsible, and ethical AI is currently a topic of general discus-sion across multiple industrial sectors and society, what this means in a manufacturing context is under-explored. Indeed, while there is a well-documented skills gap in AI within manufacturing training, this gap is even more pronounced in trustworthy and responsible AI development (UK, 2018). Hence, AI development in a manufactur-ing organisation typically one of the following two scenarios: manufacturing engineers learning how to use AI in an ad hoc manner in response to business requirements, or an AI team with little background in manufacturing tasked with implementation. Both approaches may result in risky AI practices as decisions become difficult to trace and explain. Examples include unfair bias in supplier selection, questionable surveil-lance practices around worker monitoring, failure to retrain models resulting in wrong conclusions, or erroneous maintenance predictions that lead to wasted operational "corrections" (Brintrup et al., 2022).

---

*Corresponding author. Email: ab702@cam.ac.uk

In a domain as safety critical and vital to the economy as manufacturing, there is a need to ensure the adoption of AI is both safe and appropriate, so that the envisaged societal and economic benefits are delivered responsibly. Various governments, includ-ing the UK, have urged for debate and research-informed policy and regulation in the use of AI in industry. Key to addressing these concerns, is to improve understanding of what trustworthy, responsible AI means in the domain of manufacturing and sup-ply chain management, and what particular vulnerabilities could these domains suffer from, during the AI development and deployment cycle.

In this paper we provide (i) a synthesis of the most critical issues concerning trust-worthy, ethical and responsible AI in manufacturing and supply chains under the umbrella term Trustworthy AI ; and (ii) elaborate a research agenda for future stud-ies. We outline the challenges of applying AI responsibly in manufacturing and supply chains, and explore developments in these fields that researchers in manufacturing should adopt, as well as gaps they should address. Note that our aim is not to provide a systematic literature review of definitions and aspects related to Trustworthy AI. Instead, we focus on real-world considerations in relation to the application of AI tech-nologies in the context of manufacturing and supply chains, in a trustworthy manner; our exploration is grounded on several illustrative cases drawn from current practice. In particular, we endeavour to respond to the following primary research questions:

(1) What are the specific Trustworthy AI challenges in manufacturing and supply chains?
(2) What are the research gaps in Trustworthy AI that researchers should address?

We begin by providing a focused overview of literature around Trustworthy, Re-sponsible and Ethical AI (collectively: Trustworthy AI). This is to identify common conceptualisations on the topic, emphasising a set of Trustworthy AI principles that form the basis for our analysis in a manufacturing and supply-chain context. In order to address the first question above, we use a lifecycle-based approach to illustrate how Trustworthy AI principles may be impacted, when considering the implementation of AI in manufacturing. In addition, we investigate cross-cutting considerations such as affordability and outsourcing of AI as a service. Our analysis results in 22 illustrative cases, leading to 38 research questions aimed at defining a research agenda for the manufacturing engineering and operations research community.

## 2. Definitions and Background

### 2.1. Artificial Intelligence

Definitions and interpretations of AI vary and often overlap with definitions of intelligent and autonomous systems. The original definition in the form of a 1955 Dartmouth Summer Research Project proposal (McCarthy et al., 2006) referred to AI as the "the science and engineering of making intelligent machines" and captures the essence of what AI researchers strive to achieve. In their seminal textbook on AI, Russell and Norvig (2021) have summarised the various AI definitions into four general ones, de-scribing machines that can: (1) act humanly; (2) think humanly; (3) think rationally; and (4) act rationally. Out of these, they argue that the fourth is the one that has pre-vailed throughout the history of AI, focusing on the study of agents that "do the right thing" to meet their objectives, but which, at the same time, are "provably beneficial" to humans.

In the context of manufacturing and supply chains, we adopt a definition of AI as a result of the analysis of intelligence by Legg and Hutter (2007) and Baryannis et al. (2018), focusing on two fundamental prerequisites to consider an approach as artificially intelligent: (1) the ability to autonomously decide on a course of action in order to achieve objectives; and (2) the ability to deal with a partially unknown environment. This definition admits a wide variety of approaches that include symbolic AI and knowledge representation, sub-symbolic AI, evolutionary computation and statistical AI and machine learning (ML).

Recognising the relatively increased attention that ML has received, especially in manufacturing and supply chains, in the rest of this paper we structure our presentation of trustworthy AI challenges in Section 3 primarily following the ML lifecycle proposed by Ashmore, Calinescu, and Paterson (2021). However, we make sure that descriptions of AI processes, to the extent possible, are broad enough to cover the full range of AI approaches and also add considerations that are oriented to symbolic AI, where necessary.

### 2.2. Trustworthy, Responsible and Ethical AI

Over the past decade the field of AI has experienced accelerated growth and adoption of algorithmic systems. In tandem, there is growing concern regarding the impact, implications and consequences of AI-driven systems. Some of these issues include: amplification of bias, loss of human privacy, use of AI to create digital addiction, social harms caused by digital surveillance and criminal risk assessment, disinformation through fake text generated by AI, and loss of employment or quality of employment as machines replace humans (Brundage et al., 2020). Researchers and policy makers have warned that significant efforts should be devoted to ensure the use of AI is in the public interest, that works for society and is not detrimental to humanity and human well-being. As we write this paper, there are calls from AI industry leaders themselves to embargo major AI release by six months to evaluate unintended consequences (Hern, 2023).

Much work is underway towards such concerns in a broader context, from policy makers designing regulatory frameworks to academic research proposing foundational principles for Ethical, Responsible, Trustworthy AI. This has yielded a multitude of frameworks which encourage structured, systematic exploration. Notable examples include: the assessment list for Trustworthy AI set up by the European Commission's

High-Level Expert Group on AI (High-Level Expert Group on Artificial Intelligence, 2019); the EU AI Act (European Commission, 2021); the AI Bill of Rights Blueprint by the United States White House Office of Science and Technology Policy (Office of Science and Technology Policy, 2022); and the AI Risk Management Framework by the National Institute of Science and Technology (NIST) (Trustworthy and Responsible AI Resource Center, 2023). In addition, efforts have been made to consolidate information on these issues from across different sources. For example, the NGO Algorithm Watch provides an assessment of more than 170 automated decision making systems in Europe in their 2020 Automating Society Report (Chiusi, 2020); Jobin, Ienca, and Vayena (2019) identifies 84 documents outlining different principles; and Newman (2023) introduces a taxonomy of 8 characteristics and 150 properties of trustworthiness for AI, drawn from a comprehensive analysis of the landscape of trustworthy AI. Note, however, that principles have been articulated by Western academics and technology providers, are not necessarily representative globally (Brundage et al., 2020). For example, deeper investigation showed that Beijing AI Principles show disagreements between Western and non-Western AI principles, despite them using the same terminology (Paleyes, Urma, and Lawrence, 2022). It is also worth noting that practitioners struggle with implementing these high level frameworks and regulatory guidance is missing.

While studies on the rise of AI in the manufacturing sector have been widely noted, there is no current study on the risks that it may raise in a manufacturing organisation or a set of organisations in a supply chain. Hence, much work needs to be undertaken in both interpreting principles related to trustworthy, responsible and ethical AI within a manufacturing and supply chain context, and also for ensuring that they are upheld ubiquitously.

Note that a thorough review of all existing proposals is out of the scope of this paper. Rather, we aim to briefly review some of the broader discourse around trustworthy, responsible, ethical AI as well as explainable AI to provide an indication of the common themes considered in this space, so as to provide a basis for mapping these to the challenges of AI in the context of manufacturing and supply chains. Further, for the purposes of this paper, we use the term "trustworthy AI" to refer collectively to the themes covered across the spectrum of these terms.

2.2.1. Relationships between responsibility, ethics and trustworthiness

One common approach in literature, which we follow in our analysis, is to view responsible and ethical requirements as being fundamental prerequisites to trusting an AI system. According to Smuha (2019), European Union defines trustworthy AI as being "lawful (respecting all applicable laws and regulations), ethical (respecting ethical principles and values) and robust (both from a technical perspective while taking into account its social environment)". Responsibility in the form of accountability is defined as one of seven key requirements that AI systems should meet in order to be deemed trustworthy, encompassing responsible development, deployment and use.

In his analysis of reliable, safe and trustworthy AI, Shneiderman (2020) primarily views responsibility from the perspective of clarifying the role of humans in AI failures, also mentioning responsibility in combination with fairness and explainability as goals of human-centred AI. Kaur et al. (2022) defines accountability/responsibility as one of the requirements for trustworthy AI, alongside fairness, explainability, privacy, and acceptance.

Thiebes, Lins, and Sunyaev (2021) posits that "AI is perceived as trustworthy by its users (e.g., consumers, organisations, society) when it is developed, deployed, and used in ways that not only ensure its compliance with all relevant laws and its robustness but especially its adherence to general ethical principles". For the latter, they adopt the following ethical principles: beneficence, non-maleficence, autonomy, justice, and explicability. Responsibility is only considered as an aspect of explicability and justice, in the sense of holding someone legally responsible in case of an AI failure.

Trustworthiness is also at the core of the recently published white paper of the UK Government on AI regulation (Department for Science, Innovation and Technology, 2023). One of the three main aims of the proposed regulatory framework is to increase public trust in the use and application of AI and is underpinned by five principles: safety, security and robustness, appropriate transparency and explainability, fairness, accountability and governance, and contestability and redress.

Finally, Newman (2023) also places trustworthiness at the centre of discussions around responsibility and ethics and develops a comprehensive taxonomy of 150 trust-worthiness properties. These properties relate to one of the following eight trustworthi-ness characteristics based on NIST's AI Risk Management Framework (Trustworthy and Responsible AI Resource Center, 2023): valid and reliable, safe, secure and re-silient, accountable and transparent, explainable and interpretable, privacy-enhanced, fair with harmful bias managed, and responsible practice and use. This taxonomy clearly positions responsibility and ethics as contributors and prerequisites to trust-worthiness, rather than outcomes of it.

Note, however, that an alternative viewpoint is to cast trustworthiness as a prerequi-site for responsible AI. Wang, Xiong, and Olya (2020) groups responsible AI practices into four categories: training/education, risk control, ethical design and data gover-nance. Trust building is one component of data governance, along with explainability and transparency. The narrative provided is a quite narrow view of trust that is cen-tred around reducing bias through high quality data and ensuring there is consent for sharing data.

Arrieta et al. (2020) defines seven responsible AI principles: explainability, fairness, privacy, accountability, ethics, transparency, security/safety. Trust is not included in-dependently, rather shown as an aspect or goal of explainability. As the authors ex-plain, trustworthiness and explainability are not equivalent, as being able to explain outcomes does not imply that they are trustworthy, and vice-versa.

Rather than viewing one as a prerequisite of the other, some researchers place both responsibility and trustworthiness at the same level, as principles of ethical AI. A comprehensive review of AI guidelines in literature is conducted by Jobin, Ienca, and Vayena (2019), producing the following list of ethical AI principles in order of com-monality: transparency, justice/fairness, non-maleficence, responsibility /accountabil-ity, privacy, beneficence, freedom /autonomy, trust. Responsibility and accountability are rarely defined, but recommendations focus on "acting with integrity and clarifying the attribution of responsibility and legal liability". Trust is referenced in relation to customers trusting developers and organisations and trustworthy design principles.

2.2.2. Algorithmic Ethics

Much of the discussion in this space is framed around 'ethical AI', or 'ethical algo-rithms', and thus is worth elaborating. Mittelstadt et al. (2016) developed a map with different types of ethical concerns useful for doing a rigorous diagnosis of eth-ical concerns emerging from AI, which are used to evaluate ethical outcomes in AI applications. In their map, 'inconclusive evidence' refers to the data analysis stage where results produce probabilities but also uncertain knowledge. Here authors point out cases where correlations are identified but, the existence of a causal connection cannot be posited. Failure to recognise this might then lead to unjustified actions. 'Inscrutable evidence' refers to a lack of transparency regarding both the data used to train an algorithm and a lack of interpretability of how data-points were used by an algorithm to contribute to the conclusion it generates. This is commonly referred to as the "black-box" issue leading to non-obvious connections between the data used, and resulting conclusions. 'Misguided evidence' refers to the fact that an algorithmic output can never exceed the input and thus conclusions can only be as reliable and neutral as the data they are based on, which can lead to biases. 'Unfair outcomes' refer to actions that are based on conclusive, scrutable and well-founded evidence but they have a disproportionate, disadvantageous impact on one group of people, often leading to discrimination. 'Transformative effects' refer to algorithmic activities, such as profiling the world by understanding and conceptualising it in new, unexpected ways, triggering and motivating actions based on the insights it generates (Morley et al., 2020). This can lead to challenges for autonomy and informational privacy.

Ethics by design include best practices in the development of AI to mitigate the above group of concerns – for example, the establishment of an ethics board (Leidner and Plachouras, 2017) and integration of "ethical decision routines in AI systems" (Ha-gendorff, 2020), whereby decision algorithms are explicitly designed to respect ethical values.

## 2.3. Developing Trustworthy AI

The above discussion highlights several dimensions to Trustworthy AI as described in a broader context. From an operational perspective, practitioners must consider, undertake and employ various measures and safeguards so as to mitigate the risks of the technology, such that they consider and address the various concerns to avoid negative consequences on human and societal well-being (Dignum, 2023).

Towards this, there have been various approaches describing responsible develop-ment practices. For example, Arrieta et al. (2020) and Sambasivan and Holbrook (2018) describe Responsible AI as being concerned with the design, implementation and use of ethical, transparent, and accountable AI technology in order to reduce biases, promote fairness, equality, and to help facilitate interpretability and explain-ability of outcomes.

When conceptualising Responsible AI, the principles of responsible research and innovation (RRI) have served as a starting point to consider and anticipate the conse-quences of a particular technology in society (Owen, Macnaghten, and Stilgoe, 2012). Shaped by contributions from Science and Technology Studies, this approach has been established and prominent in recent projects funded by the European Commission. The application roadmap of Responsible AI includes continuous reflection on context and civil society such as third sector organisations, so as to align the AI development pro-cess and outcomes with society's expectations. Decision processes must be visible and transparent to ensure that developers are on track regarding their responsibilities and the development process must allow users and stakeholders of technologies to criticise outcomes.

In terms of areas of consideration specific to the AI lifecycle, the topics of trans-parency (including explainability and interpretability) and fairness (bias) have received considerable attention by the technical and engineering communities, which we explore below. Note, however, that there is a clear realisation that issues of trustworthy AI are inherently socio-technical (Kroll et al., 2017; Raji et al., 2020), and require a con-sideration of technical, organisational, and human processes aspects, throughout the technology development, operation and use (Cobbe, Lee, and Singh, 2021) as well as their supply chains (Cobbe, Veale, and Singh, 2023).

### 2.3.1. Transparency

Explaining AI decisions and interpreting model outputs is commonly included in the discussion of responsible, ethical and trustworthy AI. Explainability and interpretabil-ity are often used interchangeably in literature (Arrieta et al., 2020). As argued by An-toniou, Papadakis, and Baryannis (2022), interpretability has a narrower focus that primarily relates to the degree to which ML model outputs can be interpreted in re-lation to relevant data. Explainability builds on model interpretability by including explanations that are not exclusively related to data and ML but may relate to ex-pert knowledge and other psychological, cognitive or philosophical aspects (Adadi and Berrada, 2018). An explainable approach is one that allows for identifying a complete reasoning pathway from input to output.

Three main aspects of ML interpretability are recognised in literature (De Laat, 2018). Ex-ante refers how an algorithm arrived at a decision, offered in the form of a description of the inner working of the models including what is the working procedure of an algorithm and how it generally processes input date to produce output. Ex-post refers to which training data has been used to derive results, highlighting which set of evidence / training data has been used to make each decision. A third aspect focuses on metrics used to measure the validity of the result. Here, uncertainty measures are often used, allowing users to determine confidence intervals and help them decide whether the model has made a valid decision.

It is also important to consider the role transparency plays in its broader context, raising questions about what sort of transparency, and to whom and for what. In practice, transparency generally will not solve issues with the technology (Ananny and Crawford, 2018), but can provide a basis for supporting recourse, repair, and accountability more generally (Cobbe, Lee, and Singh, 2021; Williams et al., 2022).

### 2.3.2. Fairness

Fairness refers to biases in data and deployment which can lead to systematic disad-vantages for marginalised individuals and groups. This requirement advises that AI development cycles should include methods for checking AI bias in data and decision making processes.

Many open-source ML "fairness toolkits" have been developed to assist ML practi-tioners in assessing and addressing unfairness in the ML systems they develop (Wexler et al., 2019). For instance, companies such as Microsoft, Google, and IBM, have pub-lished combinations of toolkits and guidelines that incorporate fairness. Existing fair-ness toolkits include: Fairlearn (Bird et al., 2020), AIF360 (Bellamy et al., 2019), Themis-ML (Bantilan, 2018), What-If Tool (Wexler et al., 2019), FairML (Adebayo 2016) and Fair-Test (Tramer et al., 2017).

Recent studies have shown that practitioners need more practical guidelines from fairness toolkits in order to be able to contextualise ML fairness issues and communi-cate them to non-technical colleagues (Deng et al., 2022; Lee and Singh, 2021). Deng et al. (2022) identified four design requirements ML practitioners had when using fairness toolkits: the ability to use the toolkit to learn more about ML fairness re-search, rapid use due to time constraints, ability to integrate toolkits into existing ML pipeline, and using toolkit code repositories to implement ML fairness algorithms.

2.4. Trustworthy AI Principles and their Relevance to Manufacturing

Table 1 summarises the most common requirements for Trustworthy AI proposed in literature. Although some authors use some of these terms to encompass other re-quirements, the taxonomy by Newman (2023) provides a framework that encompasses most aspects in literature.

As we see from the above terminology, there is considerable work pertaining to responsible, trustworthy and ethical AI principles and these often overlap, with debate taking place over their specific taxonomy. Defining such a taxonomy is an important area of deliberation, as it allows for structured thinking. However, for the purposes of this paper, we refrain from strictly defining this fluid field, and opt to be as broad and inclusive as possible. As mentioned earlier, we use the term Trustworthy AI as an umbrella term to encompass principles of ethical, responsible and trustworthy AI.

We argue that manufacturers need to be invested in all Trustworthy AI principles, and not just a subset of them when considering how they develop, deploy and practice AI in their organisations. Figure 1 illustrates how each principle in the taxonomy by Newman (2023) has an impact on different aspects of manufacturing.

Manufacturing companies do not exist in isolation. They impact not only the prof-itability of stakeholders but also the well-being of their workers. The products and services that manufacturing engineers design and produce impact society. The use of natural resources and waste that is generated during production and delivery have a profound impact on the environment, and decisions companies make on suppliers can have wide reaching impact on global economies. Therefore, it is crucial that manu-facturing adopts AI in a lawful, and ethical manner that is robust, safe, and avoids negative consequences to human society and well-being. It is important to be able to prove that an organisation does so, via algorithms and datasets that can be scrutinised. Given its wide ranging remit, we thus feel that the field of manufacturing needs to be inclusive when thinking about Trustworthy AI principles. Thus, in the remainder of this paper, we shall review specific challenges that manufacturing must face in order to be able to create and deploy Trustworthy AI in its broadest sense.

Table 1.: Using the taxonomy by Newman (2023), we highlight, with a few indicative examples from literature, the discrepancies that exist with regards to the definition of key requirements for a Trustworthy AI system.

| REQUIREMENTS | LITERATURE |
|---|---|
| Valid and Reliable | Mittelstadt et al. (2016) [conceptualisation], Shneiderman (2020), Kaur et al. (2022), Newman (2023) |
| Safe | Smuha (2019) [robustness, reliable, accurate, reproducible and safe], Floridi (2019) [robustness and safety], Barredo Arrieta et al. (2020) [security], Newman (2023), Shneiderman (2020) |
| Fair with Harmful Bias Managed | Mittelstadt et al. (2016), Jobin, Ienca, and Vayena (2019) [justice], Smuha (2019), Floridi (2019), Barredo Arrieta et al. (2020), Thiebes, Lins, and Sunyaev (2021) [justice], Shneiderman (2020), Kaur et al. (2022), Newman (2023) |
| Secure and Resilient | Mittelstadt et al. (2016) [manage uncertainty], Newman (2023) |
| Explainable and Interpretable | Barredo Arrieta et al. (2020), Shneiderman (2020), Wang, Xiong, and Olya (2020), Kaur et al. (2022), Newman (2023) |
| Privacy-Enhanced | Jobin, Ienca, and Vayena (2019), Smuha (2019), Floridi (2019) [privacy and data governance], Barredo Arrieta et al. (2020), Kaur et al. (2022), Newman (2023) |
| Accountable and Transparent | Jobin, Ienca, and Vayena (2019), Smuha (2019) [explainability and traceability], Floridi (2019), Barredo Arrieta et al. (2020), Wang, Xiong, and Olya (2020) [Data Governance], Kaur et al. (2022), Newman (2023), Mittelstadt et al. (2016) [risk assessment, traceability] |
| Responsible Practice and Use | Jobin, Ienca, and Vayena (2019) [inc. beneficence, non-maleficence, freedom and autonomy], Jobin, Ienca, and Vayena (2019), Smuha (2019) [lawful], Floridi (2019) [human agency and oversight, societal and environmental well-being], Barredo Arrieta et al. (2020) [ethics], Thiebes, Lins, and Sunyaev (2021)[inc. beneficence, non-maleficence, freedom and autonomy], Shneiderman (2020) [ethical], Newman (2023) |

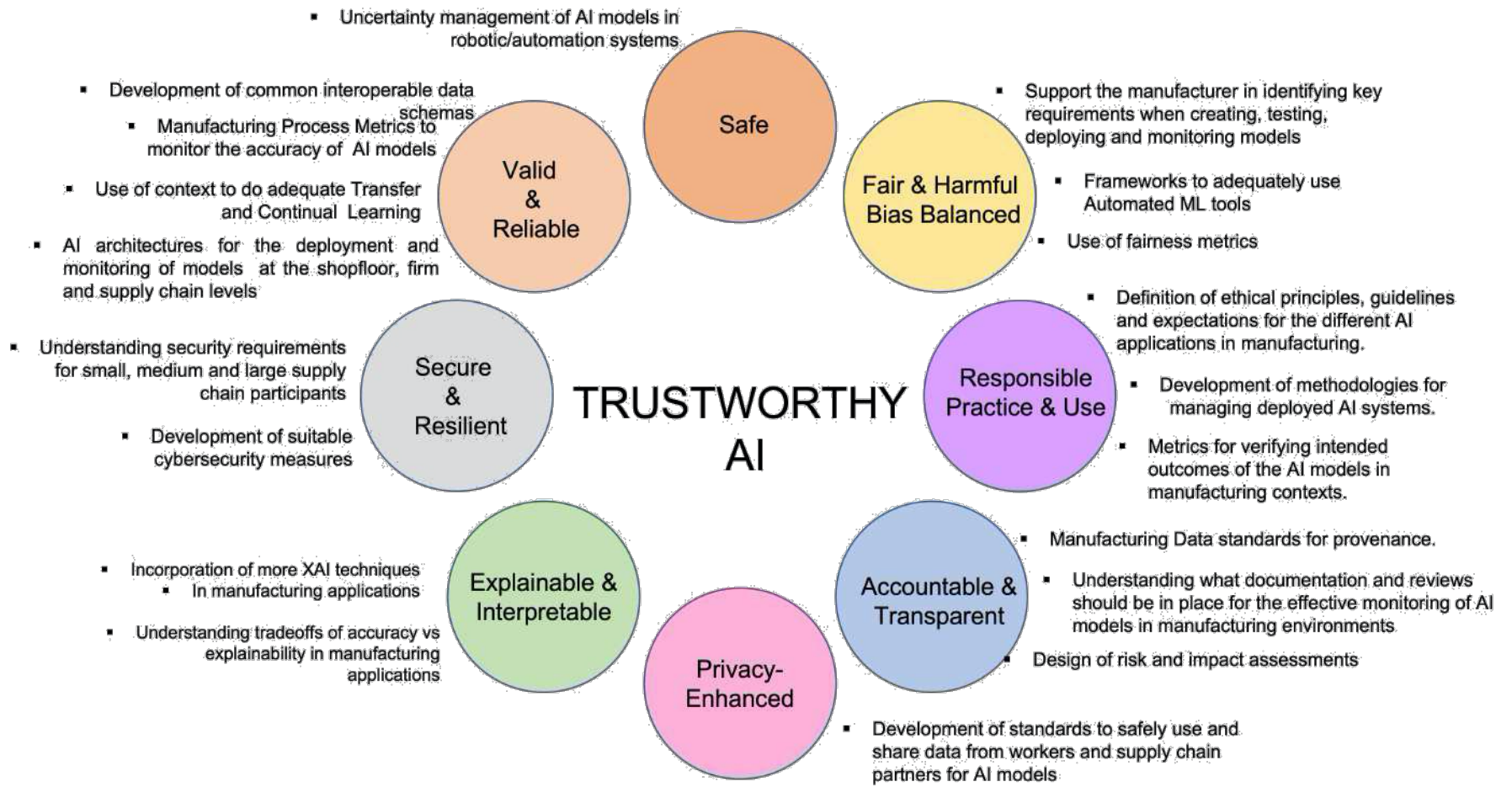

Figure 1.: Using the key principles in Newman (2023), we summarise what these mean in the manufacturing context.

3. Trustworthy AI challenges across the AI Lifecycle

The previous section discussed different, related terminology when considering trustworthiness of AI. While the frameworks associated with these concepts provide a useful starting point in understanding how AI can affect industrial contexts adversely, practitioners often highlight that they remain too general and abstract for any useful insight to be gained from them (Trocin et al., 2021; Shneiderman, 2021).

We thus propose a process-oriented lens for analysing challenges in relation to AI trustworthiness in the manufacturing context, in order to connect concerns raised in the Trustworthy AI community to the development and implementation processes that may result in their emergence. To do so, we use an adapted version of the ML life-cycle definition suggested by Ashmore, Calinescu, and Paterson (2021) and discuss, through the use of illustrative examples, how each step may result in a given trustwor-thiness concern (Table 2). While Ashmore's definition mainly encompasses statistical AI paradigms, in further subsections we will discuss the implications of Trustworthy AI in both symbolic and sub-symbolic AI approaches. To facilitate this presentation, "data" in the remainder of this section refers to both data as used in a typical statisti-cal AI approach, as well as data in the form of expert knowledge. Moreover, "model" refers to any intelligent model, ranging from ML models to knowledge models.

3.1. Data Management

Data management focuses on the preparation of datasets needed to build an AI model, which typically include data collection, augmentation and pre-processing processes.

3.1.1. Data collection

Data collection involves activities that aim to discover and understand what information is available, as well as ensuring that this information is easily accessible and processable. The task of discovering which data exists and where in an organisation is usually a challenge by itself, especially in large manufacturers with multiple facilities and geographic locations. Companies often do not know the full extent of data and knowledge they have, and what it can be used for. Data may be dispersed in emails of individuals, physical or digital documents, in legacy systems, Supervisory Control and Data Acquisition (SCADA), Manufacturing Execution Systems (MES), and Product Lifecycle Management (PLM) systems as well as structured databases such as Enter-prise Resource Planning (ERP) or Customer Relationship Management CRM) bases. In addition to these internal data, manufacturers are increasingly looking into lever-aging publicly available datasets, ranging from weather and traffic forecasts to social media.

3.1.1.1. Interoperability. When identifying different data sources relevant to the AI problem at hand, a key challenge is that these sources may have different schemas, formatting conventions, and differing storage and access requirements. For example downloading social media information may involve a different procedure than simply downloading an ERP snapshot. Joining this information into a single dataset suitable for analysis is sometimes referred to as the data integration process. Organisations may want to trial pilots using representative datasets, before setting up projects that allow easily repeatable access and integration mechanisms, which itself is determined by the usefulness of the pilot and its cost.

> Illustrative example 1: Two food manufacturers are collecting data for their products through Food Product Information Forms (FPIF), which differ in both content and structure. The manufacturers enter into an agreement that involves combining products and ingredients and need to be able to exchange data for their products and store it in a jointly managed repository. Given the differences in each manufacturer's FPIF, issues are raised with regard to data entries with different names that refer to the same concept, and data entries that are unique to one of the FPIFs. Failing to address these issues may compromise the validity of models produced based on these data.
> Risk: Missing data and wrong model, potentially leading to inscrutable evidence

Addressing data integration and interoperability issues has long been the focus of research effort in relation to information systems, especially since the proliferation of big data Kadadi et al. (2014). Standardisation through commonly agreed terminolo-gies and taxonomies is most often suggested as a solution to these issues, leveraging technologies relying on data schemas, such as those based on XML, and semantic web technologies and ontology languages, such as RDF and OWL (Pauwels, Zhang, and Lee, 2017). In this context, the following question arises:

- RQ 1: How can data producers and owners be supported in implementing common data schemas and knowledge models to improve interoperability?

3.1.1.2. Provenance. ML relies on learning models based on datasets, while knowledge-based AI relies on models built based on expert knowledge. In both cases, there is a common key challenge of identifying the sources of data and/or knowledge and of determining whether these sources can be trusted.

> Illustrative example 2: A food manufacturer is developing an automated system for determining food allergens in its food products. The system relies on a knowledge graph that is built based on food allergen information provided by other food manufacturers for each ingredient. An unreported or misreported food allergen on their part can lead to an incomplete or incorrect knowledge graph. This, in turn, may lead to unreported allergens on food product labels, which can lead to significant consequences.
> Risk: Missing or incorrect data leading to unsafe model operation, inscrutable evidence

The challenge of provenance has been well-researched in the field of knowledge engi-neering. Provenance of a resource is defined by the W3C Provenance Incubator Group (2010) as "a record that describes entities and processes involved in producing and delivering or otherwise influencing that resource". In this context, the PROV family of documents has been developed, which includes PROV-O, an ontology allowing to attach provenance information on a knowledge model. PROV-O includes three main aspects: entities, activities and agents, capturing information on agents performing activities on entities. In the standard provenance use case, this allows recording the person or company that is the source of a particular piece of knowledge.

PROV-O and related research provides the infrastructure and mechanisms to cap-ture provenance information and sets the foundations of trusting information by knowing and trusting their source. However, to ensure trustworthiness, the knowl-edge acquisition and engineering processes need to include time and effort spent on using provenance infrastructure and mechanisms to record the necessary information. This has been described in various contexts, including for AI systems (Huynh et al., 2021; Pasquier et al., 2018), both within and across their organisations and supply chains (Singh, Cobbe, and Norval, 2019), yet tends to be mostly conceptual with many opportunities for future work. This is captured in the following question:

- RQ 2: How can provenance mechanisms be leveraged in association with data and knowledge acquisition processes in a manufacturing context?

3.1.1.3. Bias. A further challenge during data collection is inadvertent introduc-tion of bias in the collected data (Suresh and Guttag, 2021). Most AI models rely on historical data to make decisions, which means if they are developed based on data that contains hidden biases, their decisions will be biased, despite the irony that many times, ML, in particular is marketed as an approach to remove human error and bias from a manufacturing use case.

Data collection issues resulting in biased data would result in inconclusive evidence by suggesting spurious correlations, unfair outcomes as data errors may result in dis-proportionate impact on one group of people or organisations, and this effect may even be difficult to detect due to inscrutable evidence, if the datasets and algorithms used are not transparent.

> Illustrative example 3: A manufacturing organisation would like to create a voice recognition system for automated robot task manipulation by shopfloor personnel. The sample that is used to train the voice recognition system will need to incorporate regional accents and have balanced gender distribution. Otherwise, female employees or employees with regional accents may not be able to use the voice recognition system, facing an unfair disadvantage compared to other employees.
> Risk: Biased training data leading to unfair outcomes

> Illustrative example 4: An automated supplier performance monitoring system is being set up to rate suppliers of a large organisation that produces engineering assets with a lifecycle of 20-25 years. Most data is collected from the organisa-tion's ERP system. However, the ERP system was implemented five years ago, and the collected dataset thus does not feature suppliers that have produced previous versions of the model, and this older data remains dispersed in indi-vidual spreadsheets. Some of the data is not accessible, as procurement officers who have developed a filing system have long retired. Thus the performance monitoring system does not contain previous data on all suppliers, resulting in bias towards newer providers.
> Risk: Missing training data leading to unfair outcomes

> Illustrative example 5: A manufacturing organisation introduces a task monitor-ing system on the shopfloor, for compliance certification and root cause analysis of any failures that occur during production. The system will monitor workers' body movements during manually intensive processes, in a bid to certify that correct process steps were followed in the right order. The data collected to train the algorithm inadvertently contained samples from male workers, whose body shape and size are different, on average, than female workers. When the system is deployed on the shopfloor, jobs undertaken by female workers are frequently flagged up as incorrect, despite it being the contrary. Risk: Biased training data leading to unfair outcomes

> Illustrative example 6: A composites producer wants to develop a worker per-formance evaluation system for manually intensive production processes. The data collected to train the algorithm initially contains the gender of the worker. Realising this feature could bias the dataset as there are much fewer women operators, the company removes the gender variable from the model. However, the analysts do not realise that the dataset contains another variable that is correlated with gender, which is the shift identifier. The female workforce tend to prefer day shifts, due to caring responsibilities. Risk: Biased training data leading to unfair outcomes

> Illustrative example 7: A consulting company wishes to estimate carbon emis-sions during a set of production processes. As carbon accounting is a manually driven, complex process, the consulting company would like to automate the estimation, by inferring carbon emissions of companies from companies that have already reported their metrics. It does so by creating a similarity measure, which takes into account company size, the sector and location it operates in, and typical production output. What is not known, however, is that the self-reported carbon emissions are incorrectly calculated in the first place, leading other companies, that have been found to be similar, to be adversely impacted from wrong estimates. Here, creating predictions from predicted data, the com-pany has confounded multiple uncertainties, yielding uninformative scores.
> Risk: Wrong data and aggregated uncertainty leading to unfair outcomes, in-scrutable evidence

The above examples pertain to challenges that may arise from biased data as well as issues with propagating uncertainties. We call for further research on the following questions:

- RQ 3: What sources of bias do manufacturing datasets and collection processes suffer from and how can they be identified and mitigated with minimal compro-mise on performance?
- RQ 4: How can informative, unbiased datasets be obtained from the shopfloor in contexts where humans are involved?
- RQ 5: How can workers who are unskilled in AI check for algorithmic bias, and fairness?
- RQ 6: How can we ensure multiple sources of uncertainty in the AI supply chain are not propagated and amplified?

3.1.2. Data augmentation and pre-processing

Data augmentation and pre-processing refer to any techniques that enhance the size and quality of datasets involved in AI approaches, particularly in the case of data-intensive approaches within ML and deep learning. This may indicatively include labelling processes to convert an unlabeled data set to a labeled one, oversampling and undersampling techniques to address data imbalance, data cleaning and feature engineering. We look at challenges related to each of these in this section.

3.1.2.1. Labeling. A label, in the context of supervised ML, is the value of an out-come variable that the ML model being developed will predict from the input data. For example, suppose one would like to predict the quality of a product from process parameters. We would need to obtain a set of data samples on products that were produced, which then relate process parameters to a quality indicator. In many real-world manufacturing scenarios, such indicators may not be easily available. Hence, one would need to label data manually. The volume of data might be too large to man-ually handle, meaning a sampling approach must be taken, where one must ensure an appropriate amount of samples are used, with appropriate variety. Alternatively, casting the problem as an unsupervised classification problem might be helpful. In addition to this, in many applications the label is subjective to the operator's exper-tise, environmental conditions or time of measurement (van Giffen, Herhausen, and Fahse, 2022). Large-scale labelling is often outsourced, potentially to non-experts or to those without sufficient domain knowledge or who will not fully aware of the intended application context, which is hard to monitor and validate and can lead to various issues downstream (Cobbe, Veale, and Singh, 2023).

> Illustrative example 8: A powder metallurgy company producing automotive parts wishes to create a quality prediction algorithm which will relate powder packing and subsequent sintering steps to resulting dimensional variance as a proxy for product quality. The production takes place in batch sizes of 1000. Due to the manual effort involved in generating the labelled data, the company opts for sampling 5 products at each process step. However, the samples during the process cannot be tracked individually, hence at each production step, different samples are taken. Because of a loss of traceability, input parameters cannot be related to the resulting quality proxy. Further, the sample size is not sufficiently representative of the variety of input parameters. The company, therefore, opts to use an unsupervised learning algorithm, to alleviate the labelling problem. Here an autoencoder approach was deployed to detect outliers as dimensional anomalies, yielding a better proxy for quality prediction.
> Risk: Small sample size, and lack of labels could result in misguided evidence

In other cases, the label of an outcome is uncertain. For instance, the expert's opinion on whether an outcome is favourable or fits into a given category might be debatable, in which case multiple experts must be consulted and a label should be agreed upon. Such instances are often the case when the ML task is a natural language processing on human generated text or speech, such as in the illustrative example that follows.

> Illustrative example 9: A study was conducted to automatically extract supply chain maps from online text data. A natural language processing methodology was used to identify companies and determine supply relationships between them. Due to the size of the dataset, Amazon Mechanical Turk was used, which is a crowd sourcing approach where human labellers are paid to annotate text data. The complexity of the task was such that labellers frequently did not agree whether a given sentence constituted a supply relationship. Thus multiple expert labellers were tasked in accordance with increased sentence complexity, and majority voting methods were used to determine the likelihood of a true label.
>
> Risk: Uncertain labels leading to wrong model, which can result in unfair outcomes, inconclusive, inscrutable or misguided evidence

In this context, majority voting is typically used. Alternatively, experts may be given different weights dependent upon past experience. Du and Ling (2010) suggest that these approaches simplify the problem by assuming uniformly-distributed noise over the sample space which fail to precisely reflect the human behaviour in real-world situations. For example, when a human is highly confident in labelling outcomes, they are naturally less likely to provide incorrect answers, whereas when such confidence is low, the noise would be more likely to be introduced. They propose "noisy label oracles" - an active learning algorithm to simultaneously explore the unlabelled data and exploit the labelled data. Peyre et al. (2017) propose weak annotations for unusual or rare labels. However, imprecise labels can lead to a loss in quality of the model making them unusable in safety critical manufacturing contexts.

3.1.2.2. Data Imbalance. In many manufacturing scenarios, the target of pre-diction is a rare event or outcome, creating a data imbalance issue. In the context of classification, data imbalance refers to cases where the positive class, i.e. the event being predicted, is by definition much rarer than the negative class i.e. an event not occurring. This may result in increased false positive rates because the biggest source of training data for the algorithm is in the majority, negative class although it is the positive class that is the main target of the predictive process. In the context of manufacturing, this bias in predictive models results in the majority of faults going unnoticed (Fathy, Jaber, and Brintrup, 2020). Data augmentation approaches that help with data imbalance include under or oversampling, or generative methods where synthetic data is generated to counterbalance the minority class. Alternatively, algo-rithmic approaches, also called "cost-sensitive learning" can be used. Here, an artificial bias is implemented in the existing classification process through a cost function that amplifies the penalty value for misclassifying minority samples.

Although class imbalance is frequently due to the nature of the data itself, at times the labelling process itself could be to blame such as in the following illustrative example.

> Illustrative example 10: An engineering company wants to predict the root cause of delays during production. They designed an interface attached onto each workstation, which asked operators to indicate the reason behind disruptions when one took place. The root causes included operator error, tool unavail-ability, machine breakdown, and random stoppage. This approach resulted in severe class imbalance, as operators almost never selected operator error, and machine breakdowns were a rare occurrence. Operators perceived data collec-tion on disruptions as time-consuming and stated that the default cause would often be random stoppage. Had the company simply used this labelled dataset, they would have misdiagnosed the reasons for delays, perhaps increasing tool buffers.
>
> Risk: Wrong training labels leading to wrong or sub-standard model, resulting in unintended consequences

For the above use cases, research questions raised in Section 3.1.1 pertaining to obtaining informative and unbiased datasets are relevant. Additionally, we ask:

- RQ 7: What are best practices to tackle imbalanced datasets in a manufacturing and supply chain context?
- RQ 8: Which methods are most appropriate for managing uncertain labels in which manufacturing contexts?
- RQ 9: How can we make sure any automated labelling done to alleviate error can still leverage the operator's expertise?

3.1.2.3. Data cleaning. Data pre-processing is commonly needed to ensure datasets are meeting requirements of the AI algorithms they are fed into. The most sig-nificant part of pre-processing and where a large amount of effort is arguably devoted is data cleaning (G´eron, 2019). This may involve identifying imputation of missing val-ues, transformation of data into a form that is applicable, and if necessary, reduction of the size of the dataset. Detection and removal of errors and decisions on whether a data point constitutes an error or an outlier, is an important aspect of the data cleaning process which, if not done properly, can result in similar issues to biased data collection, potentially yielding inconclusive evidence, unfair outcomes and inscrutable evidence.

> Illustrative example 11:
>
> A company uses goods-receipt data from one of their warehouses to predict when order will arrive, so as to optimise stock. Upon inspection, the data analytics team finds that the prediction system flags items due on Friday as three days late. Further analysis shows that the items are not late indeed, but often do not get logged onto the purchasing system until the following Monday because of reduced numbers of warehouse workers on Fridays. Had this issue not been noticed, suppliers that deliver on a Friday would have been disadvantaged as they would be categorised as low-performing suppliers.
>
> Risk: Incorrect data leading to misguided and/or inscrutable evidence, unfair outcomes

> Illustrative example 12: A train manufacturer would like to use samples with metal particulates in engine oil as a predictive feature for their prognostics algorithm, which will be used for planning maintenance. The data analytics team find out that the metal particulates for a particular train are not increasing with wear and tear as they should, but at times decreasing instead. A member of the team is sent to follow the train in operation, who finds out that the engine has an oil leak which is being topped up as it moves across the route making the underlying data irrelevant to the prediction.
> Risk: Incorrect data leading to unsafe model operation, misguided and/or inscrutable evidence

> Illustrative example 13: An analyst would like to predict product dimensions resulting from a 3D-printing process by using historical data. The analyst opts for a classification approach using a neural network but does not standardise the input features, resulting in non activation of neurons, and the result does not offer better performance than random choice.
> Risk: Lack of ML skills leading to substandard model and inscrutable evidence

3.1.2.4. Feature Engineering. As part of augmentation and pre-processing phases, it is also common to explore whether additions to variables, rather than sam-ples, is appropriate. In these cases, feature engineering is conducted, which involves creating new predictor variables from the original dataset to improve prediction capa-bility. Successful feature engineering is highly dependent on domain knowledge. Ex-perts need to agree on quantifiable hypotheses that can improve the prediction that can be extracted from the data available. Once features are created, it is important to identify whether those hypotheses were correct or not, which is influenced by the model selection as follows.

> Illustrative example 14: An example on the prediction of order delays from goods receipt data illustrates successful feature engineering. Here existing fea-tures are used to predict whether an order would be delayed. These include supplier identification, locations the product is coming from and travelling to, the product name, contractual delivery duration, and the time an order was given. In addition, one of the hypotheses put forward by the procurement team is that if a supplier is more "agile" its orders would be less likely to be de-layed. When prompted about how agility could be quantified from the existing dataset, the team designed a feature that analyses how frequently a supplier was responding positively to changing demand patterns. This feature affirmed the initial hypotheses, and led to a better predictive outcome.
> Risk: Unexploited features, leading to substandard model and inscrutable evi-dence

The above cases and discussion highlight a need to ensure domain knowledge is incorporated in the data collection effort for the prevention of errors. However, doing so should not introduce new bias as experts their own values and priorities into the context. This leads to the following research questions:

- RQ 10: What are the best methods to ensure domain knowledge is fed into AI

- RQ 11: How can we use this domain knowledge to automate model development, ensuring quality standards are met?
- RQ 12: What digital skills should manufacturing workers be equipped with so that a good synergy between the manufacturing expert and the data expert can be achieved?

### 3.2. Model creation

Following the data management part of the AI lifecycle, an AI model is created either through a model learning process, in the case of ML, or through knowledge engineering, in the case of knowledge-based AI. In this section, we focus primarily on challenges affecting model learning. This is because knowledge engineering is a human-centred process; as such, trustworthiness is less likely to be compromised as a result of the model creation process itself and is more a reflection of the trustworthiness of the human experts involved.

#### 3.2.1. Model selection

Model selection refers to selecting the type of model that will be learned. The selected ML model influences its interpretability. Failure to obtain adequate interpretability may result in inconclusive evidence and inscrutable evidence. When interpretability is prioritised the ability to interpret the output of a model plays a critical role in model selection, which then has to be balanced with computational cost as well as performance considerations. For example, decision trees (DT), which are a basic and effective ML algorithm, are widely used in practice Baryannis, Dani, and Antoniou (2019). Both Baryannis, Dani, and Antoniou (2019) and Hansson et al. (2016) de-scribe several cases in manufacturing that adopt DT because of their interpretability, ranging from supply chain risk prediction to steel production and continuous process-ing. Baryannis, Dani, and Antoniou (2019) cautions against performance loss when opting to use simpler models, and casts the model selection challenge as a trade off between performance and interpretability. It is thus important to consider multiple dimensions when it comes to model evaluation, ranging from common metric such as accuracy to fairness and interpretability. The relationship between these dimensions may not necessarily be linear, and trade offs may not be obvious. Interpretability may not necessarily mean poor outcomes, where a more interpretable or fair model might yield a small performance loss.

It is also important to note that interpretability may differ with differing model set ups even when using the same learning algorithm. For example, features that were ranked to be important in a model constructed from a dataset may not be the same features when the dataset is filtered. An illustrative example is given below.

> Illustrative example 15: The purchasing department of a manufacturing company wants to predict quotes to be received from suppliers in advance, which could then be used to detect pricing anomalies. The company collects a dataset of all of the previously ordered products, which ranges from products that are highly complex to produce, to simpler parts. A number of hypotheses are put forward by the lead procurement officer. Among them, are the effects of multi-sourcing and legacy parts. The procurement office thinks that multi-sourcing caused a deterioration in significance of individual supplier relations, hence parts that are bought from more than one supplier would be more expensive. As legacy parts are being discontinued, suppliers would think that the procurement of-fice would be "locked in" to the relationship, unable to change suppliers, hence the price would typically be higher. A price prediction model is built using a Gradient Boosted Regressor, which shows key disagreements with the hypotheses. Neither legacy parts nor multi-sourcing are significant factors, but the main factor is the supplier being asked the quote. Further analysis divides the dataset into price buckets and produces multiple models. Here the impor-tance of features shifts: for more expensive parts, it is the part complexity that was affecting price, whereas for simpler parts price is determined by supplier discretion. As the company's understanding of what drives prices grows, they are able to better focus their purchasing strategy.
> Risk: Lack of model interpretability leading to inscrutable and/or inconclusive evidence

An additional reason behind the selection of simpler models is lack of adequate computational resources, especially in resource constrained environments, where energy, memory consumption, and data transmission are limited. For example, in off-shore environments or agricultural production data transmission is limited; hence, advanced techniques, such as deep learning, are not considered yet for practical deployment, despite being able to handle high-dimensional data. Here the use of simpler models may lead to reduced performance, and hinder trust in AI algorithms. There is some work done on the development of "white-box models" from "black-box models" (Alaa and van der Schaar, 2019), by relying on symbolic (knowledge-based) AI models. However, more work needs to be done to improve the accuracy trade-off when extracting these symbolic white box models, particularly in manufacturing environments where the margin of accuracy is fundamental.

In summary, the model selection phase involves two key issues, (i) interpretability versus model performance, and (ii) the consideration of computational resources, and the environmental footprint of model training. Hence, we ask:

- RQ 13: How can we build rigorous processes to ensure the resulting outputs from ML models are explainable and interpretable in manufacturing scenarios?
- RQ 14: How can practitioners be effectively guided towards selecting the range of considerations to be prioritised for building ML models in differing contexts?

### 3.2.2. Model training

This phase involves training the chosen model with the collected, and processed dataset to learn patterns or representations of the data such that the model then can be used to cluster or classify newly observed inputs into groups, create continuous valued estima-tions about a new observation, or decide on a new action to take based on an expected value. Most ML models require hyper-parameters to be optimised during the training process such as the depth of a decision tree, the number of hidden layers in a neural network, or the number of neighbours in a K-Nearest Neighbours classifier. Finding the optimal settings of these hyper-parameters requires multiple training rounds to be run. In the worst case, the size of the hyper-parameter optimisation search space grows exponentially. Thus, as mentioned earlier, one of the biggest concerns with the model training stage is the economic cost associated with carrying out the training procedure due to the computational resources required. Strubell, Ganesh, and McCal-lum (2020) also raises an additional, growing concern around the environmental cost of training, showing that a full training cycle on a neural network could emit carbon dioxide comparable to carbon emissions of four average cars in their whole lifetime. This is especially the case when large scale language models are concerned. We depict carbon emissions of model training as a case of unintended consequences of AI, as companies are often unaware of the environmental cost of AI.

- RQ 15: How can companies make informed decisions that consider not only the cost of building ML models, but also their carbon footprint?

Another concern regarding parameter tuning is the skillset required. If the AI team are not well rehearsed in parameter tuning, the outcomes might be sub-optimal. Alternatively, automated ML (AutoML) toolkits (including online sevices) could be used, which automate various stages of the ML, including hyperparameter tuning. However, the criticism with these is that they make the process of developing ML models even more opaque, resulting in inscrutable evidence. Krauß et al. (2020) describe a use case where AutoML was pursued to predict out-of-specification products and concluded that data science expertise is necessary and cannot be replaced completely by an Au-toML system. For example, data integration, the handling of instability of training and inefficient management of the hardware resources were a challenge. Overfitting was encountered which was overcome by manual intervention. Similarly, automating various ML processes can potentially optimise for certain (often functional) aspects, while potentially ignoring aspects that might be more broadly relevant, be they around issues of bias, transparency, privacy and so on (Sun et al., 2023; Lewicki et al., 2023). For example, it has been shown that AutoML platforms might select for a user a model with the highest accuracy but which is highly biased, at the expense of models with slightly less accuracy but much reduced levels of bias (Lewicki et al., 2023). A broader point is that AutoML tooling that serves to operate generically to support anyone seeking to build a model, and therefore generally will not, nor cannot account for all the issues for the potential contexts in which the models that are automatically built will be deployed (Lewicki et al., 2023).

> Illustrative example 16: Rising energy prices increasingly necessitate more care-ful budgeting for production facilities. A machine tool producer would like to create a predictor for energy consumption at their factory using sensor based data as well as features such as production schedules and machine attributes. The company does not have a data science team and cannot afford to hire a specialist consultant. They therefore opt to use an AutoML library to automate part of the ML pipeline. The results seem promising, with over 95% accuracy and the tool is deployed to budget for energy bills. After a few months it is noticed by the accounting team that the tool vastly underestimated the en-ergy consumption, as a result of overfitting to existing datasets and inadequate training on changing production schedules.
> Risk: Lack of ML skills leading to wrong models which can cause misguided evidence. Such models may be inscrutable.

With the availability of numerous open source AI libraries, reuse of data and ML models might become increasingly commonplace, which is helpful for saving time and effort but come with no security guarantees (Gu, Dolan-Gavitt, and Garg, 2017). Code reuse, however, also creates potential security issues. One of the main privacy issues concern the preservation and leakage of the datasets collected by companies, for example through adversarial attacks that allow data reconstruction (Shokri et al., 2017) or data poisoning (Terziyan, Golovianko, and Gryshko, 2018). For example, Yampolskiy et al. (2021) illustrated how a self-learning, Internet of Things (IoT) con-nected 3D printer can be corrupted by injection of a small number of wrong labels. Researchers have developed a multitude of technical frameworks to preserve privacy during the training cycle, including explicit corruption of the data with differential privacy (Dwork, 2006), encrypted training (Gentry, 2009), and federated learning that distributes training across personal devices to preserve privacy (Zheng, Kong, and Brintrup, 2023). These are relevant to manufacturing especially in federated learning use cases and wider agent based distributed learning where datasets from multiple en-tities (Yong and Brintrup, 2020) or organisations are used to create common predictive models such as supply chain disruption prediction (Zheng, Kong, and Brintrup, 2023), industrial asset management Farahani and Monsefi (2023) and prognostics Dhada et al. (2020).

Using an interview-based methodology, Kumar et al. (2020) found that industry practitioners were not equipped with the tools to protect, detect and respond to attacks on their ML systems. The interviews revealed that security analysts either diverted responsibility to the company ML service is bought from, or expected algorithms available in commonly available platforms such as Keras, TensorFlow or PyTorch to be inherently secure against adversarial manipulations – which is not the case. The authors recommend more research to be undertaken in areas such as automated testing against adversarial attacks, threat modelling, containerisation, and rigorous forensics.

While manufacturing cybersecurity is out of the scope of this review, it is important to note that model poisoning attempts can be made by adversaries that are outside the organisation, who may gain access to ML models via IoT systems, widely deployed in industry. IBM's recent X-Force Threat Intelligence Index[1] found that the manufac-turing sector was the most attacked by ransomware, accounting for 23% of reports. Manufacturers are especially vulnerable to the algorithmic supply chain, as cyber-physical systems that are deployed are increasing. For example, industrial robots have grown from 54,000 supplied in 2010 to 121,000 in 2015, many including IoT compo-nents which pose another point of entry to industrial information systems. As robotic systems are difficult to update and deploy virus checks on due to costly downtimes, they may make AI systems vulnerable to attacks. Thus efforts to prevent data recon-struction and model poisoning should include cybersecurity checks, as illustrated in the example that follows.

---

[1] https://www.ibm.com/reports/threat-intelligence

> Illustrative example 17: A group of attackers identify a cybersecurity weakness in a manufacturer's newly installed vision recognition system used to detect objects on the work-in-progress-buffers. The system is used to automatically update the company's inventory management system by keeping track of pro-duction quantities. The adversaries implement a data poisoning attack injecting bad data into the system, causing it to misclassify objects. This is only noticed when the inventory management system gives a number of automated orders to the company's suppliers for presumably out-of-stock items, which were in fact, in stock.
>
> Risk: Adversarial attacks designed for any harmful outcomes such as unsafe operation, unethical or biased models

Adversarial attacks on models are thus a real concern, especially in infrequently updated cyber-physical manufacturing systems. We ask:

- RQ 16: What are the ways in which adversarial ML attacks can take place in manufacturing and how can they be prevented?

A growing concern is privacy of personal data used to train ML models. In a man-ufacturing scenario, this may involve end user (customer) data, as well as supplier, business relationship data, and data from employees. The ethical implications of vi-olating worker privacy are a growing concern that crucially needs more attention in the context of manufacturing. Surveillance mechanisms deployed on the factory floor are a prime example (De Cremer and Stollberger, 2022). Although manufacturing spe-cific surveys have not yet been conducted, in 2017, a global survey found that over 69% of companies with at least 10,000 employees have an HR analytics team that use automated technologies to hire, reward, and monitor employees.

In many cases, shopfloor worker monitoring may have been deployed with valid and ethical intentions. These may include, for example, ensuring Personal Protection Equipment (PPE) has been worn correctly, identifying hazards, aiding novice workers with suggestions on how to complete a difficult task, quality certification of products which necessitate an evidence trail that processing steps were performed adequately. However, the same technology can be used in ways that are ultimately detrimental to the well-being of workers, erode human-centred values and jeopardise individual rights to self-determination.

Unsurprisingly, there are limited cases that have been brought to light, and even fewer academic studies. One of the high profile cases has been reported by the Open Markets Institute, an advocacy group focussing on technology company monopo-lies (Hanley and Hubbard, 2020). They found that Amazon uses a combination of tracking software, item scanners, wristbands, thermal cameras, security cameras and recorded footage to monitor activities of warehouse workers. Whistle-blowers suggested that workers need to wear an item scanning machine (scan gun) which detected "idle time". The scan gun alerted a manager if workers spent over the maximum allowance of 18 minutes idle time per shift. Idle time included bathroom breaks, getting water, or walking slower and thus could be easily exceeded. The algorithm that powered the scan gun would classify idle time based on expected levels of motion. Two other cases from Amazon included recognising when a forklift driver has been yawning, which the drivers saw as an invasion of their privacy; and the use of employee personal data in conjunction with shopfloor worker monitoring data to prevent unionisation. In cases such as worker monitoring, there is an inherent power imbalance between the employer and employees making it hard for workers to question data being gathered about them and algorithms used to analyse their data.

Remote working during the Covid-19 pandemic has increased reports on privacy invasion, for example by software that detects worker productivity through monitoring keyboard strokes. Other commercially available software allows company managers to map company social networks by using email meta-data and detecting employee "sentiment" through email conversations, and even predicting when an employee is showing signs of frustration and may want to leave the company. The market for HR Analytics software, which include manufacturing worker surveillance, is projected to reach USD 11 billion by 2031[2].

Although worker monitoring itself is not a new concept, in the case of AI the fear is that monitoring can be scaled up by including multiple, often objectionable and pri-vate data sources (transformative outcomes), and inference is automated without any real insight to the algorithmic decision process or data itself (inscrutable and inconclu-sive evidence), potentially resulting in discriminatory practices and undue pressure on employees, effecting their well-being (unfair outcomes). The power imbalance between workers and managers means that workers often have no say or are hesitant to say whether and how such technology should be adopted.

In response, several countries are proposing regulation to prevent ethical issues arising from AI, but more research is needed to inform regulators of the potential con-sequences of the misuse of algorithms to benefit commercial interests in the manufac-turing workplace, and ensure regulation is effective. The White House Office of Science and Technology issued the Blueprint for an AI Bill of Rights in October 2022 (Office of Science and Technology Policy, 2022), with the aim of protecting civil rights and democratic values in the development and use of automated systems. The bill of rights highlights the need for data privacy in the employment context. Again in the United States (and one province in Canada), three states have implemented laws that require employers to notify employees of electronic monitoring, including AI-powered tech-nologies by providing notice to employees whose phone calls, emails, or internet usage will be monitored. In New York, automated decision tools that replace or assist in hiring or promotions decision-making must undergo annual bias audits. Companies must make the audit results publicly available, and offer alternative selection process for employees who do not want to be reviewed by such tools.

In Europe, both Norwegian and Portuguese Data Protection Authorities outlawed the practice of remote worker monitoring. The European Union is drafting an Artificial Intelligence Act (EU AI Act European Commission (2021)) to regulate AI, in which "employment, management of workers, and access to self-employment" are considered high risk and will be heavily regulated. The EU's General Data Protection Regula-tion (GDPR) limited the use of AI in employment explicitly stating that employees should not be subject to decisions "based solely on automated processing". The UK has published an AI regulation white paper (Department for Science, Innovation and

---

[2] https://www.alliedmarketresearch.com

Technology, 2023). The white paper's focus is on coordinating existing regulators such as the Competition and Markets Authority and Health and Safety Executive, but does not propose any regulatory power. Critics raised that the UK's approach has signifi-cant gaps, which could leave harms unaddressed, relative to the urgency and scale of the challenges AI bring (Hern, 2023).

The White House Office of Science and Technology issued the Blueprint for an AI Bill of Rights in October 2022, with the aim of protecting civil rights and democratic values in the development and use of automated systems. The bill of rights highlights the need for data privacy in the employment context. Again in the United States (and one province in Canada), three states have implemented laws that require employ-ers to notify employees of electronic monitoring, including AI-powered technologies by providing notice to employees whose phone calls, emails, or internet usage will be monitored. In New York, automated decision tools that replace or assist in hiring or promotions decision-making must undergo annual bias audits. Companies must make the audit results publicly available, and offer alternative selection process for employ-ees who do not want to be reviewed by such tools. Both Norwegian and Portuguese Data Protection Authorities outlawed the practice of remote worker monitoring. The European Union is drafting an Artificial Intelligence Act (AI Act) to regulate AI, in which "employment, management of workers, and access to self-employment" are considered high risk and will be heavily regulated. The EU's General Data Protection Regulation (GDPR) limited the use of AI in employment explicitly stating that em-ployees should not be subject to decisions "based solely on automated processing". The UK has published an AI regulation whitepaper (Department for Science, Innovation and Technology, 2023). The whitepaper's focus is on coordinating existing regulators such as the Competition and Markets Authority and Health and Safety Executive, but does not propose any regulatory power. Critics raised that the UK's approach has significant gaps, which could leave harms unaddressed, relative to the urgency and scale of the challenges AI bring (Hern, 2023).

While the regulatory debate is encouraging, we note that the effectiveness of these regulatory initiatives in the manufacturing and supply chain sector remains to be seen. At the moment these frameworks are not enforceable, and progress is not fast enough to keep up with the fast pace of AI research developments. Although multiple high level frameworks have been created, as summarised in Section 2.2, there is a lack of practical use cases and guidance on how these frameworks can be incorporated to daily business practice. This is especially true in the field of manufacturing.

At the time of writing, ISO, IEC and BSI standards are being discussed and pro-posed [3] (e.g. ISO/IEC TR 24028:2020) However, at the moment there is no consen-sus on the adoption of trustworthy AI standards. Should companies wish to develop surveillance mechanisms on their workers to prevent, for example unionisation, they are free to do so. Participation in industry standards may play an important role to design effective regulatory frameworks. The emergent interplay between standards de-velopment and regulatory approaches will be a decisive factor, and multiple, clashing standards and regulation might stifle progress in the area. Regulatory enforcement of standards may be criticised for stifling innovation, whereas too laissez-faire an ap-proach may yield unintended consequences as discussed in this section. NGOs such as the Algorithmic Justice League already helped scrutinise and remove bias from a number of facial recognition algorithms used by the police force in the US. Similar ap-proaches can be taken in manufacturing. Researchers also suggested independent audit firms could develop reviewing strategies for AI projects and make recommendations to their client companies about what improvements to make. The insurance industry could also help guarantee trustworthiness by specifying requirements for underwriting AI systems in manufacturing.

---
[3] https://aistandardshub.org/

We therefore call for more research in understanding how manufacturing organisations exploit AI technology in ways that can breach human privacy and well-being, and what mitigation mechanisms and guidance can be designed to prevent such breaches:

- RQ 17: Should we build a manufacturing sector specific code of conduct that interprets and adapts existing legal instruments pertaining to the use of AI?
- RQ 18: How will we ensure interoperability between the various AI regula-tory frameworks and standards that are currently being developed in different regions?
- RQ 19: How will multinational manufacturing organisations adopt differing standards in their supply chains?

> Illustrative example 18: A vision recognition algorithm has been designed for a factory shopfloor during Covid-19 for ensuring workers followed social distanc-ing rules and wore PPE. The system would warn employees when distancing rules were not met. The system was designed to be private, and would hold no personally identifying information, only using generic object recognition. How-ever, leaked documents and subsequent media interviews with workers have suggested that the system was combined with personal data, such as number complaints raised and used for an additional purpose: prevention of unionisa-tion. The company used additional datasets on worker background and personal information to estimate a risk score for unionisation. Based on the score, high risk individuals would be warned to keep distance from certain individuals or reallocated to different shifts.
>
> Risk: Unethical surveillance or misappropriation of data leading to privacy vi-olation, unfair and/or transformative outcomes

> Illustrative example 19: A supply chain surveillance algorithm is deployed to help improve supply chain visibility at a company. The tool will improve an understanding of which geographic locations is the company's upstream sup-pliers are concentrated on, so that risk mitigation measures can be taken. The tool predicts a link between one of the company's suppliers and an "anonymous firm". Although the firm name is anonymous, because of the additional infor-mation revealed, including geographic location and production, the company infers that the supplier has also been selling to their main competitor.
>
> Risk: Unethical surveillance or misappropriation of data leading to privacy vi-olation, unfair and/or transformative outcomes

Brundage et al. (2020) identified intitutional governance to be another key candidate for improving ethical practices. They suggest visible leadership commitment includ-ing regular review board meetings, annual, publicly available responsible AI reports, and reward mechanisms for responsible AI practices can be valuable in increasing incentives in organisations. Brundage et al. (2020) also suggest inter-institutional re-porting mechanisms such as NASA's Aviation Safety Reporting System and the Food and Drug Administration's Adverse Event Reporting System and Bugzilla as useful models for technical reporting.

As seen above, a large number of privacy and ethical challenges can arise during the model training phase. Data collection and model training are intertwined when it comes to trustworthiness. Whilst in Section 3.1.1 we highlighted that data needs to be bias free and obtained under consensus, and be able to be scrutinised by the owners and generators of data who might be unskilled in AI, in this section we raise additional questions pertaining to personal data that is used to train ML models. These include:

- RQ 20: What is the definition of personal data in a manufacturing context? How can workers know how their data is used and can they have a right to consent or decline the way their personal data is used?
- RQ 21: Should policy makers build mechanisms to ensure that data that was originally collected for its purpose remains its purpose in a manufacturing con-text?
- RQ 22: What robust institutional mechanisms can be put in place to empower employees to scrutinise AI models in an organisation? What methods can be used by unskilled workforce to check for algorithmic bias and fairness, as well as decision processes that impact them, which rely on AI?
- RQ 23: Counterarguments on surveillance activities on suppliers and workers highlight that surveillance has always been practiced, and the only difference AI brings is scale and accuracy. Does the manufacturing community, including generators of personal data agree with this statement? In a manufacturing con-text, can we achieve consensus on what types of data collection and algorithmic surveillance constitute fair and unfair outcomes?
- RQ 24: For unethical practices, how can the right balance between regulation and innovation be found? Should specific AI standards for manufacturing be built and if so, should they be enforceable in different contexts?

## 3.3. Verification

Verification of AI models should include rigorous checks to ensure they are robust and reliable in satisfying functional and performance requirements. Here, robustness refers to the degree to which the developed model can function correctly in the pres-ence of invalid inputs or varying environmental conditions, while reliability refers to the probability that the model performs required functions for the desired period of time without failure. Verified artificial intelligence has been defined as "AI-based systems that have strong, ideally provable, assurances of correctness with respect to mathematically-specified requirements" (Seshia, Sadigh, and Sastry, 2022).

In the case of knowledge-based models, verification is rooted in the strong mathe-matical logic based foundations of such models and leverages extensive research and development efforts in model checking (Clarke Jr et al., 2018) and automated theo-rem proving (Bibel, 2013), as well as formal specification languages (Baryannis and Plexousakis, 2013, 2014; Baryannis, Kritikos, and Plexousakis, 2017). For instance, ontology-based knowledge models can be verified through an array of established rea-soning systems, such as HermiT (Glimm et al., 2014) or Pellet (Sirin et al., 2007), available through established, user-friendly tools, such as Protégé[4].

In contrast, verification of neural network based ML models is a hard problem due to their black-box nature, making approaches such as the aforementioned model checking or theorem proving, or even source code reviews not applicable (Salay, Queiroz, and Czarnecki, 2017). One of the issues is the large size of the state-space, making the design of test cases difficult. Reinforcement learning (RL) approaches especially suffer from large state-spaces, as the decision space is non-deterministic and the system might be continuously learning meaning that over time, there may be several output signals for each input signal.

---

[4] https://protege.stanford.edu/

Automated test case generation (Clark et al., 2014), transfer learning and synthetic data generation (Borg et al., 2018) have been proposed as potential solutions. El Mhamdi, Guerraoui, and Rouault (2017) suggested the robustness of a deep neu-ral network could be evaluated by focusing on individual neurons as units of failure. Continuous monitoring of model input (elaborated further in Section 3.4), and inte-grating verification processes across the whole development cycle rather than at the end (for example by experimenting how output varies in the state-space with different model architectures) have been proposed as best practice (Taylor, 2006). Adler, Feth, and Schneider (2016) proposed coding a "safety-cage", where the model execution is turned off with increased uncertainty and swapped with a deterministic model track. Although these suggestions stem from the field of autonomous systems, notably au-tonomous vehicles, they are worth noting and re-interpretation within the context of manufacturing is worth exploring.

Test based verification usually involves the setting up of a simulation based test environment, which is typically safer, cheaper, and faster to run. However, as with any simulation based methodology, conclusions derived is dependent and constrained by the assumptions made by the simulation designer. Even small discrepancies between the simulation environment and real world can cause dramatically different outcomes, exemplified by high-profile cases in the field of autonomous vehicles (Dulac-Arnold et al., 2021). Focusing on RL, the authors highlight several challenges with the trans-fer of RL algorithms from simulation based training to real-life environments. These include limited sample size, delays in task rewards, constraints, unexpected, stochastic changes in environment, multiple objective functions. Stochasticity means that agents are not guaranteed not to explore unsafe conditions, which may have unintended conse-quences, unless these are thought by the designer in advance and coded into the reward function, which is often infeasible for the designer to capture exactly what we want an agent to do. Consider a robot tasked to pick up items in a warehouse from delivery zones and place them in designated locations. The algorithm designer may simply code a reward function to maximise the number of items picked and placed. In the simula-tion the robotic agent works in a fairly constrained, stable environment. In reality, the layout of the warehouse may change, with moving obstacles which may include human workers. If the agent has not been trained to avoid moving obstacles, and its reward is based on number of tasks completed, it could explore taking unsafe shortcuts (called reward-hacking). While RL shows much promise as a learning paradigm, its imple-mentations in real-world settings remain very limited (Wang and Hong, 2020). As RL is starting to be popularised in manufacturing robotics (Oliff et al., 2020), condition-based maintenance (Yousefi, Tsianikas, and Coit, 2020), vehicle-routing (Mak et al., 2021), and inventory control (Kosasih and Brintrup, 2022; Wang and Hong, 2020), it is worth exploring these challenges in real-world manufacturing environments.

> Illustrative example 20: A company would like to implement autonomous clean-ing robots in their warehouse to speed up operations after a shift ends. The company has bought ML-as-a-service from a well known AI solution provider. The provider has multiple success stories with warehousing giving confidence to the purchasing company on its credentials. In a bid to speed up the train-ing process they use transfer learning, which involves extrapolating a reward function for the new warehouse environment based on reward functions from other similar cleaning robot algorithms they have developed. The provider also sets up a period of observation in the warehouse to ensure this approach would work in the new setting. For a period the pilot seems successful. Following a change in the cleaning materials used, an incident happens leading to a fire as the robot follows an unsafe shortcut where a chemical process is taking place as it was not explicitly coded not to do so, which should have been the case after transfer learning.
> Risk: Insufficient verification leading to unreliable, unsafe operation

ML performance metrics should be carefully considered by the application con-text. While the accuracy of a classifier is a well-known and widely used metric, in safety critical applications such as machine failure other metrics may be more ap-propriate (Baryannis, Dani, and Antoniou, 2019). For instance, recall may be more important, which measures the number of correctly identified positive classes over all classes that should have been identified as positive. In other cases, where incorrect identification of a class is costly, precision may be used – for example when a pro-duction or quality delay is falsely predicted, resulting in inventory build up. Other considerations may include quantifying the uncertainty of predictions, both in a clas-sification and a regression context. Hence, performance metrics should be carefully designed and reflect contextual priorities. Additionally, performance checks should in-clude checking for bias and fairness to overcome some of the issues relating to unfair outcomes, discussed in previous sections. Here inherent bias in the training data can be checked by comparing ranges of features to the actual distribution of the feature in the real world across different data slices.

> Illustrative example 21: Consider a manufacturer developing a classifier for predicting supplier delays, which will then be used to optimise buffers. The manufacturer may choose to optimise the training cycle using precision or re-call. Precision refers to the ratio of correctly predicted delayed-orders over all delayed-order predictions, and recall refers to the ratio of correctly predicted delayed-orders over the number of actual delayed-orders. False classification of an on-time order could lead to unnecessary risk mitigation actions, such as building inventory buffers that might be costly. On the other hand, false classi-fication of a delayed-order as low risk could be more problematic, as the costs of dealing with an unexpected disruption could outweigh mitigation planning. The manufacturer may need to weigh these objectives for example using an F-measure, which allows one to combine these two objectives and weigh each one differently. Risk: Incorrect model objectives leading to unintended consequences

The main concerns at this phase include safety and reliability of the developed models in noisy manufacturing contexts:

- RQ 25: How can established approaches in knowledge-based AI verification, such as model checking be leveraged in the case of ML?
- RQ 26: How can test case generation methods developed be applied to manu-facturing use cases?
- RQ 27: How can stochasticity of manufacturing environments be captured in ML test environments in a meaningful way?

Further, during the training phase, performance metrics can have varying impact on safety, cost and efficiency. We ask:

- RQ 28: How do different performance metrics effect outcomes in differing man-ufacturing contexts?
- RQ 29: How can insurance and/or legal coverage be ensured for continuously adopting AI models?

3.4. Model deployment

Model deployment refers to the operationalisation of the model that has been built, by building the software infrastructure that is necessary to run it, and setting and following policies on model maintenance and updates. One of the major challenges in this stage is identifying when a model needs to be updated with new information, and to what extent older information should continue to be utilised. In the case of ML models, "concept drift" describes the situation where the feature distribution shifts over time due to a change in the underlying data-generating process. Concept drift means that the mapping between features and output no longer matches the new incoming data. Thus, as real-world contexts evolve and adapt to changes over time, the underlying datasets that are representative of the system should change. For example, a demand forecasting model used to predict demand for a fashion product should be retrained frequently. However, in other cases the changes in the system may not be contextually obvious to the model owner, in which case input data should be continuously monitored.

The two main ways to deal with concept drift are to update the model incrementally, or to retrain the whole model, which can be done periodically, based on predefined performance criteria (such as accuracy or F1 scores), or statistical approaches based on uncertainty quantification (such as Hoeffding bound).

> Illustrative example 22: A forging machine is equipped with a condition monitoring algorithm to estimate time to failure, based on the number and desired shape of parts that it handles. Whilst this is initially successful in reducing unplanned downtime by correctly estimating service needs, overtime the accuracy of predictions decrease. Upon inspection, it is found that the reason has been that the material specification of the main batch produced by the machine has slightly changed along with the supplier of the raw material, creating higher loads on the machine.
> Risk: Concept drift leading to misguided evidence

Through this illustrative example, it is made clear that the model deployment phase is continuous. Challenges resulting from this involve finding the right frequency and method of model updating and detecting when a model is no longer applicable to the context it is deployed in:

- RQ 30: Which applications in manufacturing are prone to concept drift? Which drift detection methods are more informative in manufacturing?
- RQ 31: What are the best practice mechanisms to identify models used in manufacturing that are no longer useful and should be updated or decommissioned?

4. Cross-cutting considerations

In this section we briefly discuss cross-cutting Trustworthy AI challenges that manufacturing organisations face when considering the adoption of AI technology within their organisations or across their supply chains.

4.1. Affordability

A key issue in adopting AI technologies is cost, which not only includes time and effort spent across the development and deployment steps of AI, but also cost of data access, storage and post-deployment costs such as maintenance. Studies performed on the adoption of digital manufacturing technologies, which include AI, show that adoption is often contingent upon affordability. In the UK, for example, over 99% of businesses are small to medium enterprises (SMEs, 0 to 249 employees[5]) with lower affordability, which might create a larger capability discrepancy in supply chains. It is worth noting that many SMEs are vital to supply chains of larger organisations, hence the success of the manufacturing industry is intertwined. The manufacturing community needs to monitor and encourage SME adoption and we propose the following research questions in this context:

- RQ 32: How can SMEs access benefits of AI solutions?
- RQ 33: Does affordability of AI impact Trustworthy AI adversely?

4.2. Outsourcing AI as a service

While no current statistics exist on the extent to which AI is developed in-house versus bought as-a-service, both approaches are not without challenges for manufacturers, and outsourcing decisions are likely to depend on a number of factors including the specificity of development, longevity of its use, AI skills the company would like to retain, the degree of control a company wants to exert on the algorithmic approaches developed and external infrastructure dependencies. Whilst classical theories such as Resource Based View and Transaction Cost Economics may offer useful starting points for framing AI outsourcing decisions, additional considerations on the trustworthiness of algorithms may need to be factored in. As mentioned earlier, outsourcing AI may make it more difficult to investigate bias in data used to train algorithms, as well as affect algorithmic safety. Moreover, it might be tempting to outsource in an attempt to try and shift accountability and responsibilities elsewhere, which raises legal and broader governance considerations (Cobbe, Veale, and Singh, 2023; Cobbe and Singh, 2021) We call for more research on AI outsourcing decisions:

- RQ 34: Are existing decision frameworks for outsourcing applicable for AI as a service in manufacturing and, if not, how should they be extended?

---

[5] https://www.gov.uk/government/statistics/business-population-estimates-2022

- RQ 35: How can Trustworthy AI be ensured when outsourcing AI in manufacturing?

## 4.3. Data compensation and monetisation

Data is one of the most valuable assets of firms that fuel much of the digital econ-omy today. Often datasets are traded amongst companies by so-called data brokers. As a society, we often do not know how and where our data is used, and for what purpose, forming an active field of ethical and regulatory debate. In manufacturing scenarios, companies may use data not only from individuals (such as monitoring how a consumer uses their products, or workers producing them), but also from other companies (such as monitoring activities of their suppliers and competitors). At the moment, generators of these datasets are typically not compensated. Whilst this con-stitutes an ethical issue, in other cases data compensation may open up new markets and opportunity. For example, manufacturers may be interested in tapping into other companies' datasets which they lack. For example, if a manufacturer would like to implement prognostics for its machinery but does not have enough run-to-failure data it may be able to appropriate datasets from another manufacturer using the same machine. The data owner may wish to monetise its datasets. The automotive industry have been discussing how customers can be compensated by sharing car usage data so they can design better (McKinsey and Company, 2020). Data compensation is a strongly debated field where no regulatory guidance currently exists. Researchers have proposed the use of blockchain for tracking how data is used and develop mechanisms to compensate owners of data (Maher, Khan, and Prikshat, 2023). We call for more research on how manufacturers can monetise their datasets and also compensate data owners ethically and responsibly:

- RQ 36: Which types of manufacturing data can be monetised?
- RQ 37: Would monetising data have an effect on AI trustworthiness in manu-facturing and what are the associated risks and legal implications?

## 4.4. Scalability of Trustworthy AI

Although some large manufacturers and supply chain businesses have made some advances in the incorporation of AI-based solutions into their decision-making and process control, for most companies (particularly SMEs) AI systems remain at the pilot stage (a situation often referred to as "pilot purgatory"). These pilot studies, as is the case with most AI solutions, implement models tailored to a specific use case and developed with data that manufacturers are typically not keen to share. However, as seen in the success of AI in other sectors such as finance or e-commerce, the secret to scalability and production-ready models is in sharing these models and data across businesses. Hence, to achieve the full potential of AI in the manufacturing value chain, it is recognised that models and data need to be transparent and usable but at the same time secure to protect intellectual property and privacy (Davis et al., 2022). More research and development is needed to better understand the positive and negative effects from provenance to privacy and finding best ways to develop and share models from aggregated data without losing sight of security issues:

- RQ 38: How can Trustworthy AI solutions be made more scalable in manufac-turing?

5. Conclusions and Managerial Implications

AI can be a significant driver in improving productivity in manufacturing and supply chains, but it also can be misused with unsafe, unethical practices. Whilst in the Western World there is progress towards a set of commonly agreed principles, there is significant confusion on what they mean in practical terms. As AI technology is moving rapidly in manufacturing, there is an urgent need to guide manufacturers on the risks that come with AI adoption and deployment, so that its benefits can be delivered safely and ethically.

In this paper, we have conducted a brief review of terminology in the field of Trust-worthy AI, after which we mapped potential risks that arise during the AI development and deployment lifecycle, using illustrative use cases. This thought exercise has shown that Trustworthy AI risks may be present throughout the complete AI lifecycle, from data collection to post-deployment, as summarised in Table 2. We also highlighted a number of cross-cutting concerns that may be present throughout the entire AI devel-opment process. Doing so yielded 38 research questions aimed at guiding research into Trustworthy AI in manufacturing and supply chains. In addition to guiding research, we hope that the mapping provided will help practitioners identify the types of risks they should pay attention to whilst they go through stages of AI development in their organisation.

The research questions that have been raised can be classified into three main cat-egories: risks pertaining to data collection and processing, algorithmic development and deployment, and organisational practice of AI in manufacturing.

Risks pertaining to data collection, and processing: Incorrect, misrepresented or historically biased data may all be present in manufacturing AI use cases. Our analysis showed that manufacturing is at risk of both specific instances of data collection bias (for example shopfloor personnel labelling why errors occur in production), but also of discriminatory bias, such as the use of datasets with inherent bias on gender or ethnicity. Aggregation of bias is also a risk factor, as uncertain labels or measurement are used to train models. Another common problem in manufacturing stems from the target of predicting rare events such as machine failures or supply delays, which yield data imbalance. Organisations need to adopt skills to scrutinise bias in datasets and remove them. Researchers need to create both technical advances in identifying and removing bias, and correlate types of bias in datasets with the manufacturing scenarios that yield them. Simplifying bias exploration and empowering employees who are unskilled in AI with the tools to scrutinise datasets is another important gap that researchers may wish to focus on.

Risks pertaining to algorithmic development and deployment: Algorithmic risks are those that stem from inappropriate or insufficient design and use of AI algorithms. Researchers have highlighted the trade offs between explainable and interpretable al-gorithms and performance, as well as noting the often large carbon footprint of model training. Future lines of research should consider the exploration of novel approaches, such as neurosymbolic AI (Garcez and Lamb, 2023), that have increased explainabil-ity and are operator-centred, to ensure these provide the intended decision support. Guidelines need to be developed to detect concept drift in a variety of representative manufacturing scenarios, and to decide between model updating and de-commissioning of models that are no longer useful. Model verification in realistically designed envi-ronments that capture the stochasticity of manufacturing scenarios, knowledge-based AI verification, and test case generation are further areas that are in need of attention. Open source algorithms and public datasets can accelerate research in this area.

Table 2.: Summary of Trustworthy AI challenges in manufacturing

| Process | Illustrative cases | Research questions | Issue | Trustworthy AI Challenge | Trustworthy AI Principles | Potential directions |
|---|---|---|---|---|---|---|
| Data collection | 1 | 1 | Missing data, wrong model | Inscrutable evidence | Valid & Reliable | Develop common interoperable data schemas |
| Data collection | 2 | 2 | Missing or incorrect data | Inscrutable evidence, unsafe operation | Valid & Reliable, Safe, Accountable & Transparent | Leverage provenance mechanisms |
| Data collection | 3-6 | 3-5 | Biased or missing data | Unfair outcomes | Fair with Harmful Bias Managed | Check data bias using fairness toolkits, Establish ethics board |
| Data collection | 7 | 6 | Wrong data, aggregated uncertainty | Unfair outcomes, inscrutable evidence | Fair with Harmful Bias Managed, Explainable & Interpretable | Check data validity |
| Data augmentation | 8 | 7 | Small sample size, lack of labels | Misguided evidence | Valid & Reliable, Responsible Practice and Use | Unsupervised learning, Incorporate domain knowledge |
| Data augmentation | 9 | 8 | Uncertain labels | Unfair outcomes, inconclusive, misguided, inscrutable evidence | Valid & Reliable, Fair with Harmful Bias Managed, Explainable & Interpretable | Revise data augmentation process, Incorporate domain knowledge |
| Data augmentation | 10 | 9 | Wrong labels | Unintended consequences | Valid & Reliable | Crowd sourced labelling, Majority voting, Noisy oracles, Weak annotations |
| Data pre-processing | 11-12 | 10-11 | Incorrect data | Misguided, inscrutable evidence, unsafe operation | Safe, Valid & Reliable | Check data validity |
| Data pre-processing | 13-14 | 12 | Lack of ML skills | Inscrutable evidence | Valid & Reliable, Responsible Practice and Use | ML expert input |
| Model selection | 15 | 13-14 | Explainability | Inscrutable, inconclusive evidence | Explainable & Interpretable, Accountable & Transparent | Explore explainable AI methods, Domain expert input |
| Model training | 16 | 14 | Lack of ML skills | Misguided evidence | Valid & Reliable, Safe, Responsible Practice and Use | ML expert input |
| Model training | 17 | 16 | Adversarial attacks | Any harmful outcome, such as unsafe operation | Secure & Resilient | Adopt cyber security measures |
| Model training | 18-19 | 17-24 | Unethical surveillance, Data misappropriation | Privacy, unfair and transformative outcomes | Privacy-Enhanced | Regulation, Adherence to standards, Establish ethics board |
| Verification | 20 | 25-27 | Insufficient verification | Unreliable, unsafe operation | Valid & Reliable, Safe | Automated test case generation, Safety cages, Synthetic data, Transfer learning |
| Verification | 21 | 28-29 | Incorrect objectives | Unintended consequences | Valid & Reliable, Accountable & Transparent, Responsible Practice and Use | Use of appropriate performance metrics |
| Model deployment | 22 | 30-31 | Concept drift | Misguided evidence | Valid & Reliable, Secure & Resilient, Responsible Practice and Use | Adopt appropriate model monitoring and update processes |

Risks pertaining to failures in Trustworthy AI Practice: Much technical research needs to be undertaken to identify and remove biases in data, interpret algorithmic results, verify models before deploying them, detect when an algorithm needs to be updated or de-commissioned. However, implementation of these technical advances are dependent on organisational practices that not only allow but also encourage such scrutiny and corrections to take place. We raised several research questions that are in need of attention relating to the right balance between regulation and innovation, and practical implementation of regulatory frameworks or standards. There is currently no clear consensus on the types of data collection and algorithmic surveillance that would constitute unfair outcomes. A further complicating factor for manufacturers whose supply chains span across multiple jurisdictions is potentially differing AI standards. The outsourcing of parts of the AI lifecycle may yield further issues where responsibility of development and verification is dispersed and perhaps untraceable. Organisations need to develop robust mechanisms to ensure safety from adversarial attacks, and for improving ethical practices. Wide reaching, visible leadership commitment should be considered with practices appropriate to the organisational setting, such as ethical review boards, publicly available responsible AI reports, whistleblowing roles, and reward mechanisms to increase incentives for Trustworthy AI practice.

It is also worth noting that while this paper presents the first comprehensive fram-ing of Trustworthy AI in manufacturing, and highlights the risks associated with AI, it represents only a first step in this increasingly important area. Both the field of AI, and AI adoption in manufacturing are rapidly evolving, as do the definitions of what constitutes Trustworthy AI. As we have seen, terminology pertaining to Ethics, Trust-worthiness, and Responsible AI are not yet agreed upon, although there is emergent consensus on the main high-level principles such as fairness, accountability, safety, and responsible use that is beneficial to humans. Researchers working at the intersection of AI and manufacturing engineering need to not only familiarise themselves with these developments but also take an active role in shaping and interpreting them for man-ufacturing. More illustrative use cases and best practices need to be brought to light, in order to guide research-informed practice.

6. Data Availability statement

Data sharing not applicable – no new data generated

7. List of Figures and Alt-Text

Figure 1 Caption: Using the key principles in Newman (2023), we summarise what these mean in the manufacturing context
Figure 1 Alt text: Summary of Trustworthy AI principles in Manufacturing